\documentclass[sigconf]{aamas}

\usepackage{balance} 
\usepackage[inline]{enumitem}
\usepackage{tikz}
\usetikzlibrary{positioning}
\usepackage[ruled]{algorithm2e}
\usepackage{amsmath}
\usepackage{dsfont}
\usepackage{dblfloatfix}
\usepackage{bbm}

\setcopyright{ifaamas}
\acmConference[AAMAS '22]{Proc.\@ of the 21st International Conference
on Autonomous Agents and Multiagent Systems (AAMAS 2022)}{May 9--13, 2022}
{Online}{P.~Faliszewski, V.~Mascardi, C.~Pelachaud,
M.E.~Taylor (eds.)}
\copyrightyear{2022}
\acmYear{2022}
\acmDOI{}
\acmPrice{}
\acmISBN{}

\acmSubmissionID{154}

\title{Socially Supervised Representation Learning: the Role of Subjectivity in Learning Efficient Representations}

\author{Julius Taylor}
\affiliation{
  \institution{Inria - Flowers Team\\
  Université de Bordeaux }
  \city{Bordeaux}
  \country{France}}
\email{julius.taylor@inria.fr}

\author{Eleni Nisioti}
\affiliation{
  \institution{Inria - Flowers Team\\
  Université de Bordeaux \\
  ENSTA ParisTech}
  \city{Bordeaux}
  \country{France}}
 \email{eleni.nisioti@inria.fr}

\author{Clément Moulin-Frier}
\affiliation{
  \institution{Inria - Flowers Team\\
  Université de Bordeaux\\
  ENSTA ParisTech}
  \city{Bordeaux}
  \country{France}}
 \email{clement.moulin-frier@inria.fr}

\begin{abstract}
Despite its rise as a prominent solution to the data inefficiency of today's machine learning models,  self-supervised learning has yet to be studied from a purely multi-agent perspective. In this work, we propose that aligning internal subjective representations, which naturally arise in a multi-agent setup where agents receive partial observations of the same underlying environmental state, can lead to more data-efficient representations. We propose that multi-agent environments, where agents do not have access to the observations of others but can communicate within a limited range, guarantees a common context that can be leveraged in individual representation learning. The reason is that subjective observations necessarily refer to the same subset of the underlying environmental states and that communication about these states can freely offer a supervised signal. To highlight the importance of communication, we refer to our setting as \textit{socially supervised representation learning}.  We present a minimal architecture comprised of a population of autoencoders, where we define loss functions, capturing different aspects of effective communication, and examine their effect on the learned representations. We show that our proposed architecture allows the emergence of aligned representations. The subjectivity introduced by presenting agents with distinct perspectives of the environment state contributes to learning abstract representations that outperform those learned by a single autoencoder and a population of autoencoders, presented with identical perspectives of the environment state. Altogether, our results demonstrate how communication from subjective perspectives can lead to the acquisition of more abstract representations in multi-agent systems, opening promising perspectives for future research at the intersection of representation learning and emergent communication.
\end{abstract}

\keywords{Representation Learning; Emergent Communication; Data Augmentation; Multi-Agent Systems}

\newcommand{\BibTeX}{\rm B\kern-.05em{\sc i\kern-.025em b}\kern-.08em\TeX}

\begin{document}


\pagestyle{fancy}
\fancyhead{}


\maketitle 


\section{Introduction}
\label{sec:intro}
Since their recent introduction in problems with large search spaces and complex dynamics, machine learning (ML) models no longer have the luxury of tabula rasa training. Upon encountering a new task, the modern ML practitioner fine-tunes an existing model that has been pre-trained on a large set of tasks, motivated by the observation that priors learned during the pretraining phase will ensure improved data efficiency during the finetuning phase. Crucially, the pretraining phase takes place in the absence of supervised signals, endowing the agent with the responsibility of extracting a supervisory signal from unlabelled observations by leveraging some known structure of the data, a paradigm termed as self-supervised learning \citep{bert,schwarzerPretrainingRepresentationsDataEfficient2021, grill2020bootstrap}.

\begin{figure}[b]
    \centering
    \includegraphics{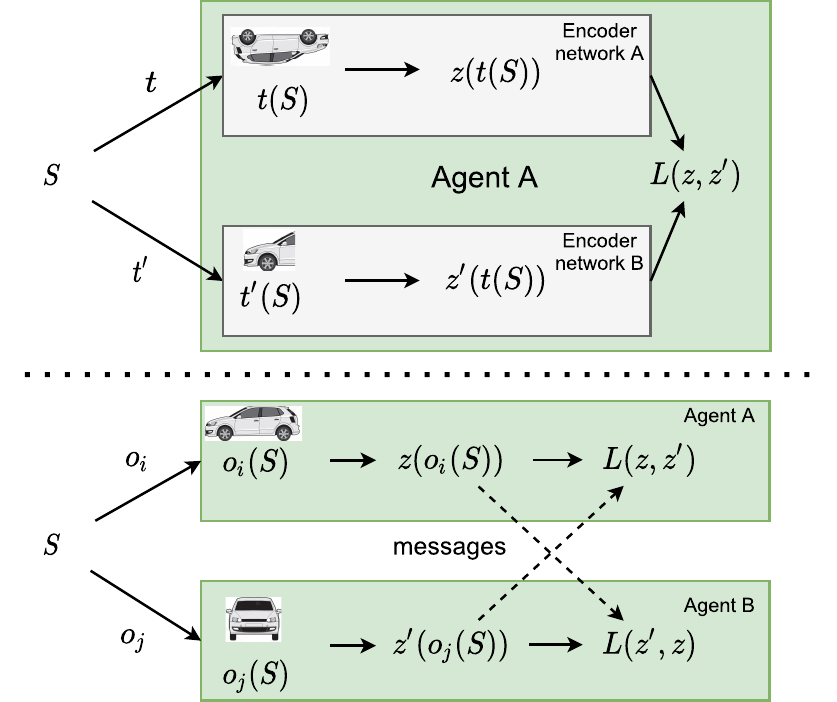}
    \caption{Data augmentation in self-supervised representation learning using stochastic image augmentations $t \sim T$ (top) vs. Socially Supervised Representation Learning that substitutes engineered augmentations for perspectives ($o_i, o_j, \dots$) that arise naturally in multi-agent systems (bottom). Green boxes indicate conceptual \textit{agents}, while we assume that a singular representation learning method may be interpreted as a single agent.}
    \label{fig:my_label}
\end{figure}

Designing a self-supervised agent requires an application-specific understanding of which aspects of the data's structure can provide a meaningful and strong supervisory signal. Prominent approaches in this area are: 
\begin{enumerate*}
\item building world models that allow the agent to collect fictitious experience and avoid continuous expensive interaction with the environment \citep{atari, world_models};
\item learning representations that optimise unsupervised objectives, such as exploration \citep{liu2021unsupervised} or contrastive-learning losses \citep{srinivasCURLContrastiveUnsupervised2020, chenSimpleFrameworkContrastive2020a,grill2020bootstrap};
\item learning predictive representations of some aspect of an agent's internal state such as velocity \citep{shangAgentCentricRepresentationsMultiAgent2021}.
\end{enumerate*}

An important technique that has been combined with these approaches is that of \textit{data augmentation}, which relies on artificially augmenting samples from the dataset through a set of predefined transformations. These transformations are meant to preserve the semantic characteristics of the original examples by leading to representations invariant to local changes in the input, which have been observed to offer better generalization and data efficiency  \citep{vonkuegelgen2021selfsupervised}. In vision tasks, transformations are traditionally geometric manipulations of an original image, such as translation, rotation and zoom, as they preserve the content of an item while changing non-semantic aspects. The self-supervised paradigm has lead to significant advancements in many data-hungry applications, such as Natural Language Processing \citep{bert}, Atari games \citep{atari} and robotic manipulation \citep{zhan2020framework}. Little attention has been, however given to multi-agent settings, where the exponential complexity increase exacerbates data inefficiencies. In this work, we formulate and answer a novel, exploratory question: ''Does the multi-agent setting confer opportunities for learning data-efficient representations in an unsupervised manner beyond the mere transfer of techniques developed in single-agent settings?''

Our work attempts to answer this question by formulating a multi-agent architecture that leverages the data augmentation paradigm in a shared environment. As an illustrative example of a real-world or simulated setting, we can envision a group of embodied agents navigating in an environment populated by a variety of objects. We assume that agents have access to a communication channel and do not impose any restriction on the type of downstream tasks they will solve after the unsupervised pretraining phase under study. Three central observations of our work in such a setting are:
\begin{enumerate*}[label=(\roman*)]
\item The full state of the environment is not directly observable to the agents. Even when observing the same object, and thus perceiving the same subset of the full state space, two agents may be receiving different observations -- depending, among others, on their angle, distance to the object and sensor characteristics. As these observations are internal to an agent and therefore reflect their subjective ``point of view'', we refer to them as \textit{perspectives}; 
\item The grounded nature of the environment offers a free guarantee that observations received simultaneously and immediately communicated about refer to the same underlying state;
\item Under these conditions, multiple observations from different point of views are analogous to different transformations used in data augmentation, under the condition that they only communicate in close range.  
\end{enumerate*}

Based on these observations, we propose a new learning paradigm that we call \emph{Socially Supervised Representation Learning} (SocSRL). SocSRL leverages the similarities between data-augmentation in self-supervised learning and multi-agent subjective observations in a shared environment. We hypothesise that \textit{subjectivity} in multi-agent systems provides "for free" the same benefits as data augmentation in single-agent representation learning: Multiple agents observing different perspectives of the same underlying environment state are analogous to the application of multiple transformations in data augmentation. Disparities among the different agents' observations incentivise an alignment of internal representations towards a more "objective" representation that abstracts away the particularities of the subjective inputs. However, our framework differs from classical, single-agent approaches to data augmentation in some important respects: 
\begin{enumerate*}[label=(\roman*)]
\item the transformations that produce the augmented views are not manually designed but naturally emerge from multiple agents partially observing a shared environment;
\item each agent is a black box to other agents, meaning that they cannot access their respective inputs and the parameters of their internal representations (in contrast to the majority of works in single-agent data augmentation \citep{grill2020bootstrap, chenSimpleFrameworkContrastive2020a, srinivasCURLContrastiveUnsupervised2020});
\item agents can only share information by communicating about the internal representation of their current subjective observation.
\end{enumerate*}

In this paper, we test this hypothesis in a simplified setting. We use labelled datasets (MNIST and CIFAR-10) as a simplified model of a multi-agent environment. We make the following analogies between a multi-agent environment and a labelled dataset: (1) The true environment state, hidden to the agents, is analogous to the class label in the dataset (e.g. the digit class in MNIST). (2) Different subjective observations of the same underlying environment state by different agents are analogous to different samples of the same underlying class label in the dataset.

\paragraph{\textbf{Contributions}.} We summarise our contributions as follows:
\begin{enumerate}[label=(\roman*)]
    \item We highlight an interesting link between data-augmentation traditionally used in single-agent self-supervised setting and a group of agents interacting in a shared environment.
    
    \item We introduce \textit{Socially Supervised Representation Learning}, a new learning paradigm for unsupervised learning of efficient representations in a multi-agent setup.
    
    \item We present a detailed analysis of the conditions ensuring both the learning of efficient individual representations and the alignment of those representations across the agent population.
\end{enumerate}

\section{Related work}
\label{sec:related_work}


The overarching benefit of effective representations is that of capturing task-agnostic priors of the world. Capturing such abstract properties has been equated with learning representations invariant to local changes in the input \citep{bengio_represent}. \citet{NIPS2009_428fca9b}, for example, find that deep autoencoders exhibit this ability in image recognition tasks. However, representations used by an encoder-decoder pair result from the co-adaptation of the two networks and are, therefore, not abstract in any objective sense. In contrastive learning \citep{srinivasCURLContrastiveUnsupervised2020, chenSimpleFrameworkContrastive2020a}, positive examples, generated through data augmentation, are used to pull together content-wise similar examples in the latent space, and negative examples are pulled apart. While contrastive learning requires researchers to manually define criteria for forming positive and negative pairs, as well as data augmentation functions, our work uses the inherent properties of multi-agent systems to naturally infer positive samples.


Aligning the representations learned by a group of agents has been approached from different perspectives:
\begin{enumerate*}[label=(\roman*)]
\item the common practise of sharing a single network that each agent feeds with its own observations \citep{Gupta2017CooperativeMC} trivially avoids the problem of a potential misalignment. However, it requires that agents have the same observation space and share a common internal model, which is an unrealistic assumption in most multi-agent settings;
\item in emergent communication setups \citep{lowePitfallsMeasuringEmergent2019a,foersterLearningCommunicateSolve2016}, representations are aligned through communication, but a supervised signal, on the form of a truth function or a shared reward, is required to infer communicative success. In contrast, our setting only requires that agents receive subjective observations of the same underlying environment state.
\end{enumerate*}

Community-based autoencoders \citep{tieleman_shaping_2019} also attempt to leverage a multi-agent setup to improve learned representations. Here, a population of encoders and decoders randomly forms pairs, which forces them to learn more abstract representations by avoiding the aforementioned co-adaptation effect. Although closely related to ours, this work trains all autoencoders on the same data and does not exploit the subjectivity inherent to multi-agent systems.

\section{Methods}
\label{sec:methods}
\begin{figure*}
    \centering
    \includegraphics{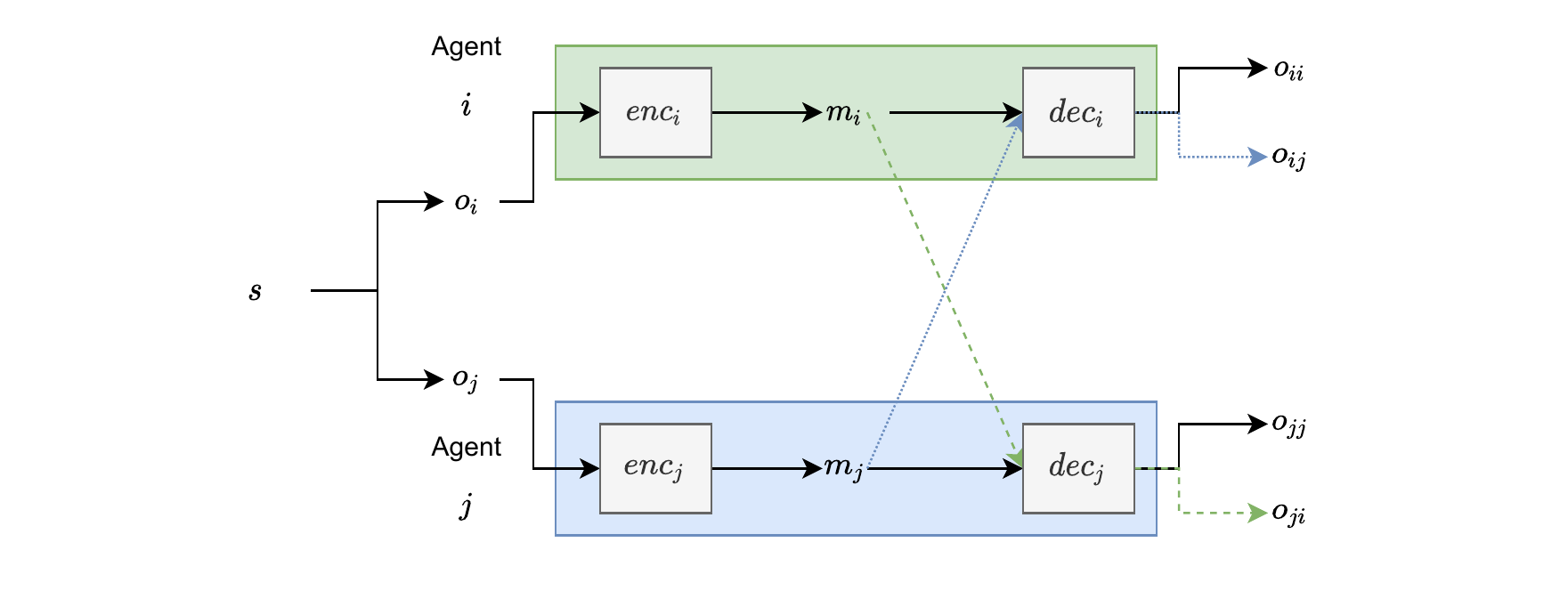}
    \caption{Proposed architecture. Two agents $i$ and $j$ (top and bottom coloured boxes) are presented with different observations ($o_i$ and $o_j$) of the same underlying environment state ($s$). Each agent (say $i$) implements a standard autoencoder architecture with an encoder $enc_i$ mapping the input observations to latent representations and a decoder $dec_i$ mapping latent representations to reconstructed observations. Each agent $i$ communicates its latent representation $m_i$ (also called a message) of its own observation input $o_i$ to the other agents. This way, each agent $i$ is able to reconstruct the observation from its own latent representation ($o_{ii}$: reconstruction by agent $i$ from its own message $m_i$) as well as from the latent representation of the other ($o_{ij}$: reconstruction by agent $i$ from the other's message $m_j$). The architecture can be trivially extended to a larger population, where each agent communicates their latent representations to with each others.}
    \label{fig:agent_arch}
\end{figure*}

\subsection{Problem definition}
We consider a population of agents $\mathcal{A}$ and environment states
$s \in \mathcal{S}$, hidden to the agents. Each agent $i\in \mathcal{A}$ receives a private observation of
the state $o_i(s)\in \mathcal{O}$, where $\mathcal{O}$ is an observation space. Agents are essentially convolutional autoencoders, though other self-supervised learning technique could be used (for example variational autoencoders \citep{kingma2013auto}). We define encoder and decoder functions $\text{enc}_i: \mathcal{O}\rightarrow \mathcal{M}$ and $\text{dec}_i: \mathcal{M}\rightarrow \mathcal{O}$, respectively, where $\mathcal{M}$ is a latent representation space (also called a message space, see below). Given an input observation, an agent $i$ encodes it into a latent representation $ m_i := enc_i(o_i)$ and attempts at reconstructing the observation through $o_{ii} := \text{dec}_i(m_i)$, dropping the dependence on $s$ for brevity. Agents will use these latent vectors to communicate to other agents about their perceptual inputs (hence the term message space for $\mathcal{M}$). When agent $i$ receives a message from agent $j$ they decode the message using their own decoder, i.e. $o_{ij} :=  \text{dec}_i(m_j)$. The diagram in Fig.~\ref{fig:agent_arch} depicts a possible setup with 2 agents.

Importantly, while the observations provided to the agents at each time step are sampled from the same environment state, we systematically ensure that agents never access this state, nor the input observations and the reconstructions of each other. This makes our approach applicable to a wide range of decentralized multi-agent settings.

Given this architecture and a dataset mapping each state $s\in \mathcal{S}$ to a set of observations in $\mathcal{O}^{N_s}$, where $N_s$ is the number of observations available in the dataset for state $s$, we are interested in the following research questions:
\begin{itemize}
    \item Under which conditions can the agents converge towards aligned representations? (see below for the definition of alignment measures)
    \item Does SocSRL improve the efficiency of the learned representations compared to a single agent baseline? If so, what are the main factors influencing it?
\end{itemize}
\subsection{Losses for communication}
\label{sec:losses-for-comm}
In order to incentivise communication in our system, we define four loss functions which encourage agents to converge on a common protocol in their latent spaces. With $MSE$ being the mean squared error, we define the message-to-message loss as
\[ L^{MTM} = \text{MSE}(m_i, m_j), i \neq j.\]
This loss directly incentivises that two messages (i.e. encodings) are similar. Since messages are always received in a shared context, this loss encourages agents to find a common representation for the observed state, abstracting away particularities induced by the specific viewpoint of an agent. Next, we propose the decoding-to-input loss, given by
\[L^{DTI} = \text{MSE}(o_{ij}, o_i), i\neq j.\]
This loss brings the decoding of agent $i$ from agent $j$'s message closer to agent $i$'s input observation, indirectly incentivising an alginment of representations because both agents can reconstruct from the other agents message more easily, when they agree on a common latent code i.e. they have similar representations for a given $S$. Then, we propose the decoding-to-decoding loss:
\[L^{DTD} = \text{MSE}(o_{ii}, o_{ij}), i\neq j,\]
which is computed using the reconstructed input of agent $i$ and the reconstruction of $i$ incurred from the message sent by $j$. Lastly, the standard autoencoding loss is given by
\[
L^{AE} = \text{MSE}(o_{i}, o_{ii}).
\]

Especially when optimising for $L^{MTM}$, there exists a potential Achilles' heel in our system: the trivial solution of mapping all data points to a single point in latent space, yielding $L^{MTM} = 0$. Thus, optimising $L^{MTM}$ will always require to be optimised in conjunction with losses which prevent this degenerate solution, such as $L^{AE}$. We further add noise to the communication channel to a) simulate a more realistic communication environment and b) to enforce more diverse representation in the latent space to prevent it from collapsing when optimising $L^{MTM}$. The message of agent $i$ is thus defined as $m_i = \text{enc}_i(o_i) + \epsilon$, with $\epsilon \sim \mathcal{N}(0, \sigma)$, where $\sigma$ is a hyperparameter in our system. The total loss we optimise is thus \[L = \eta_{MTM} L^{MTM} + \eta_{DTI} L^{DTI} + \eta_{AE} L^{AE} +\eta_{DTD} L^{DTD}\] with $\eta_{MTM}$, $\eta_{DTI}$, $\eta_{AE}$, and $\eta_{DTD}$ being tunable hyperparameters.

We analyse our methods in a setting of \emph{differentiable communication}, we allow gradients to pass the communication channel barrier during optimisation (see e.g. \citet{foersterLearningCommunicateDeep2016a} for a similar setting in emergent communication). This means that gradients are backpropagated from the decoder of an agent (e.g. $dec_j$ in Fig.~\ref{fig:agent_arch}) to the message sent by another agent (e.g. $m_i$ in Fig.~\ref{fig:agent_arch}).

\subsection{Environment}
We study our method in a simplified setting where we emulate multi-agent subjectivity with labelled datasets and use independent samples from a randomly chosen class as substitute for perspectives. At each step, we determine the state $S$ by drawing from a discrete uniform distribution, i.e. $ S \sim \mathcal{U}\{0, \rho - 1\}$, where $\rho$ is the number of different classes in our dataset. We then compute a subset of our dataset, $\text{DATASET}(S)$, containing only samples of the class $S$. Finally, we derive agents' individual observations by sampling a batch of data from $\text{DATASET}(S)$ uniformly at random without replacement. Since our datasets do not contain duplicates, we can ensure that agents receive pairwise distinct samples of the currently active class $S$.

To make sure that our method is not dataset-dependent, we run our experiments on two datasets with varying complexity. We use MNIST (GPL license) \citep{deng2012mnist} and CIFAR \citep{krizhevsky2009learning} (MIT license), which both contain 60000 images of 10 classes in total.
\subsection{Training procedure}
To train our system, we sample two agents without replacement $i, j$ from the population of agents $\mathcal{A}$ and then minimise the losses we defined in section~\ref{sec:losses-for-comm}. For each round of multi-agent training, we also train a pair of regular autoencoders with no access to multi-agent losses (only $L^{AE}$) and use these agents as baseline for comparison. The full algorithmic loop is described in the Algorithm~\ref{alg:ma-ae}.
\begin{algorithm}
    \SetAlgoLined
    Initialise a population of agents $\mathcal{A}$\\
    Dataset $\mathcal{D}$ with $\mathcal{C}$ classes\\
    \While{not converged}{
        Sample two agents $ i, j \in \mathcal{A}, i\neq j $\\
        Sample random class  $s \sim \mathcal{U}\{0, |\mathcal{C}|-1\} $\\
        Sample inputs $ o_{i}, o_{j} \sim \mathcal{D}(s) $\\
        Agents $i$ and $j$ compute and minimise loss
        $ L = \eta_{AE} * L^{AE} + \eta_{MTM} * L^{MTM} + \eta_{DTI} * L^{DTI} + \eta_{DTD} * L^{DTD}$\\
    }
    \caption{Socially-supervised representation learning}
    \label{alg:ma-ae}
\end{algorithm}

\subsection{Evaluation}
\label{sec:evaluation}
We want to measure if the latent representations learned by agents are able to capture useful features for downstream tasks, as well as if they converge to aligned representations across the agent population. We consider that agents have learned aligned representations when similar inputs (i.e. images of the same class) are encoded by similar messages across the agent population. Thus, we introduce measures for quantifying these properties. First, in order to check whether the representations capture important properties about the data, for each agent $i$, we train linear probes $f_i(\boldsymbol m)$ on an evaluation set to predict class identity from the learned latent space, following the commonly used procedure described in \cite{grill2020bootstrap, chenSimpleFrameworkContrastive2020a, tieleman_shaping_2019, kolesnikov2019revisiting, kornblith2019better, oord2018representation}. If representations are able to capture informative properties of the data, this classification task will be easier, thus we consider classification accuracy as a proxy for the quality of representation. When training linear classifiers, we always freeze the agents weights such that only the linear part is learned. 

Curiously, works like \citet{henaff2020data} and \citet{resnick2019probing} show that learning simple linear probes might not be sufficient to assess if representation capture interesting properties of the data. To that end, we take inspiration from \citet{whitney2020evaluating} and evaluate the data-efficiency of our representations in addition to classification performance. For that, we ask the question of how many data points we need to adapt to a downstream task using our representations. To answer this question, we generate $K$ random subsets of our validation set, $\mathcal{V}_i \subset V$, where $V$ is the validation set, $|V_i| = 10^{x_i}$, and $x = \{r: r = 1 + n \times \frac{\log_{10}|V|}{K - 1}, \quad n \in \{0, \dots, K - 1\}\}$, by sampling the $v \in \mathcal{V}_i$ from $V$ uniformly at random without replacement. This is equivalent to creating random subsets of $V$ whose magnitudes are spaced evenly on a $\log_{10}$ scale. We conduct our experiments with $K = 10$, such that for the smallest set we have $|\mathcal{V}_0| = 10$ and for the biggest set we have $|\mathcal{V}_9| = 10000 = |V|$. We evaluate our learned representations on all $\mathcal{V}_i$ by further splitting the $\mathcal{V}_i$ into a randomly chosen training set $\mathcal{V}_i^t$ and validation set $\mathcal{V}_i^v$ with an $80:20$ ratio and record the performance of the linear probes $f$ on all $\mathcal{V}_i^v$ after training them on $\mathcal{V}_i^t$. If our representations are data efficient, i.e. they need fewer new data points to adapt to downstream tasks, performance gains over the baseline will be more pronounced for smaller $\mathcal{V}_i$.

Next, in order to measure the alignment of representations, each agent uses a trained linear classifier (which in this case was trained on a randomly generated $\mathcal{V}_9$) and classifies latent representations of the dataset computed by \textit{other agents}, thus computing $f_i(m_j)$ with $i \neq j$. We record the zero-shot classification performance (without any additional training), which we call \textit{swap accuracy}. Further, we define an additional proxy for alignment, called the \textit{agreement}, which is computed as
\[
    \frac{1}{{|\mathcal{A}|}^2 - |\mathcal{A}|}\frac{1}{|D|}\sum_{d \in \mathcal{D}}\left[
    \sum_{i\in \mathcal{A}}
    \sum_{j\in \mathcal{A}, i\neq j}
    \mathbbm{1}_{f_i(enc_j(d)) = f_j(enc_i(d))}
    \right],
\]
and can be described as the fraction of time agent $i$ and $j$ agree on the same class label, when using the encoder of agent $i$, $enc_i$, as input. With $d \in \mathcal{D}$ we denote a sample from the dataset.

\subsection{Hyperparameter search}
\label{sec:hp_search}
To determine the best combination of hyperparameters, we run a search over the space of possible combinations. In order to balance exhaustiveness, flexibility, and computation time, we opt to use a mixture of grid search and random sampling. For each hyperparameter controlling a loss term, i.e. $\{\eta_{AE}, \eta_{MTM}, \eta_{DTI}, \eta_{DTD}\}$, we uniformly sample a value between 0 and 1. We then evaluate this combination of losses at four different values of noise levels, ie. $\sigma \in \{0, 0.33, 0.67, 1.0\}$. Additionally, to inspect each loss in isolation, we run a special set of runs where one specific $\eta$ is set to 1, and the others to 0. Those runs are also evaluated at the aforementioned noise levels.


\section{Experiments and results}
\label{sec:experiments_and_results}
\begin{table}[H]
    \caption{Excerpt from our hyperparameter analysis showing the top 2 performing runs (DTI, AE+MTM) and a baseline (AE). Sorted by accuracy.}
    \label{tab:hparam-results}
    \centering
        \begin{tabular}{lllllllr}
        \toprule
         $\eta_{AE}$ & $\eta_{MTM}$ & $\eta_{DTI}$ & $\eta_{DTD}$  &  Test accuracy (\%)  & Alias\\
        \midrule
         0.0 &     0.0 &     1.0 &     0.0 &    \textbf{95.1}  & DTI\\
          0.81 &    0.14 &    0.03 &    0.01 &   \textbf{92.3} & AE+MTM\\
          \vdots \\
          1.0 &     0.0 &     0.0 &     0.0 &   87.3 & AE\\
          \vdots\\
           0.0 &       1.0 &       0.0 &       0.0  &  11.4 & MTM \\

        \bottomrule
        \end{tabular}
\end{table}
\begin{table}[H]
    \caption{p-value annotation legend.}
    \label{tab:annotation_legend}
    \centering
    \begin{tabular}{lc}
\toprule
symbol & p \\
\midrule
ns & 0.05 $<$ p $\leq$ 1\\
* & 0.01 $<$ p $\leq$ 0.05\\
** & 0.001 $<$ p $\leq$ 0.01\\
*** & 0.0001 $<$ p $\leq$ 0.001\\
**** & p $\leq$ 0.0001\\
\bottomrule
\end{tabular}
\end{table}
\begin{figure}
    \centering
    \includegraphics{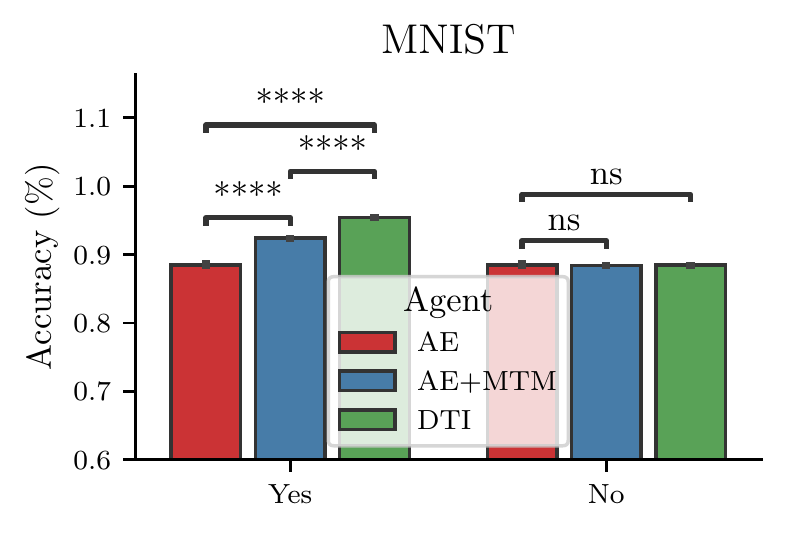}
    \includegraphics{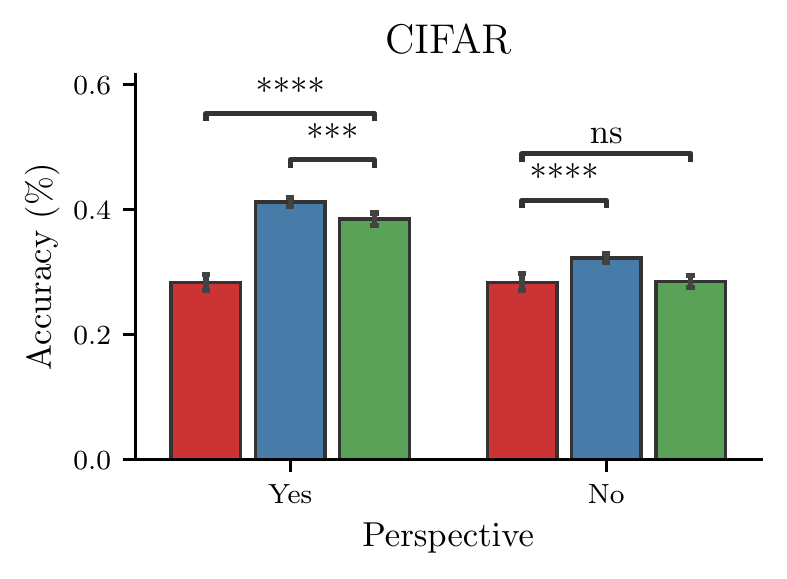}
    \caption{Classification accuracy when predicting class identity from the learned latent space, evaluated for different variables, using \textit{differentiable communication}. Left: Accuracy for different runs and with or without perspective. Right: Accuracy of best run (DTI) as a function of the number of agents per population.}
    \label{fig:classification-accuracy-mnist}
\end{figure}
If not stated otherwise, all experiments are run with ten unique random seeds and error bars indicate the 95\% confidence interval (bar plots) or the standard deviation (line plots). Significance is determined using an unequal variances t-test (Welch's t-test). We annotate significance levels in bar plots using the symbols defined in Table~\ref{tab:annotation_legend}. We use the statannot package\footnote{https://github.com/webermarcolivier/statannot} for annotation (MIT license). We will report results both obtained with the MNIST dataset and the CIFAR10 dataset in the following sections.

Before measurements, we train populations of 3 agents (if not stated otherwise) for 50000 total rounds according to the procedure described in Algorithm~\ref{alg:ma-ae}, using a batch size of 1028. We use a lab-internal, distributed architecture which allows the training on multiple Nvidia V100 GPUs concurrently. In total, we consumed roughly 50000 GPU hours for all experiments.

We analyse 400 parameter settings in each condition and rank each individual experiment according to downstream classification accuracy. In the following sections, we will focus on three specific parameter configurations, two of which performed the best in our parameter sweep (\textit{DTI} and \textit{AE+MTM}) and one autoencoding baseline (\textit{AE}). These configurations are further described in an excerpt of this parameter sweep in Table~\ref{tab:hparam-results}. Against our expectations, we found that $L^{DTD}$ was the only loss that did not positively impact downstream classification performance. Additionally, we find that the best results are achieved at an overall noise level of $\sigma = \frac{2}{3}$.

To our surprise, we found no well-performing parameter setting which included a significant $\epsilon_{DTD}$. Our interpretation is that solving for $L^{DTD}$ is prone to generate a degenerate solution, where decoders simply learn an input-independent mapping to an arbitrary vector, effectively reducing $L^{DTD}$ to 0.

\subsection{Performance evaluation of the learned representations}
\label{sec:mnist}

We first investigate the quality of the learned representations and what factors contribute to increased downstream performance. To that end, we train populations of agents and then predict class identity from the learned latent space using a linear classifier, as described in section \ref{sec:methods}. 
We additionally conduct an ablation study where we supply all agents from the population with the same input images, instead of using different images from the same underlying class.  

We display the results for MNIST in Fig.~\ref{fig:classification-accuracy-mnist} (top), which show that DTI yields the highest classification performance. In fact, DTI finds representations which result in classification performance significantly better compared to the baseline (AE) and AE+MTM, whereas AE+MTM significantly outperforms AE. When removing the perspectives and supplying all agents with the same input images (Figure~\ref{fig:classification-accuracy-mnist} bar group on the right), our methods do not provide a significant improvement above the baseline anymore. 

For CIFAR10, we show plots in Fig.~\ref{fig:classification-accuracy-mnist} (bottom). Here, we observe that while \textit{DTI} and \textit{AE+MTM} both significantly outperform the baseline, \textit{AT+MTM} significantly outperforms \textit{DTI}. As with MNIST, we also find that ablating the perspectives from our systems is detrimental to performance, albeit the effect is not as strong as with MNIST. Here, we still observe a significant improvement for \textit{AE+MTM}, whereas \textit{DTI} remains at baseline performance.

Both results on MNIST and CIFAR10 show that our method is able to learn improved representations compared to autoencoding baselines and that varying perspectives among agents are crucial for performance gains. While \textit{DTI} seems to be a strong choice for simple datasets like MNIST, where we find low inter-class variability, this setting does not seem to transfer without loss to a more complex dataset like CIFAR10. Nonetheless, aligning latent spaces via our proposed $L^{MTM}$ loss yields significant improvements both in MNIST and CIFAR10 and thus presents itself as a robust and effective way to exploit the subjectivity emergent in multi-agent interaction. Note that \textit{AE} \textit{with perspective} and \textit{AE} \textit{without perspective} are equivalent, as perspectives only affect scenarios where our multi-agent losses are optimised, which are set to 0 in \textit{AE}. Therefore, we can infer that a statistically significant improvement is gained in \textit{AE+MTM without perspective} versus \textit{AE with perspective}.

\begin{figure}
    \centering
    \includegraphics{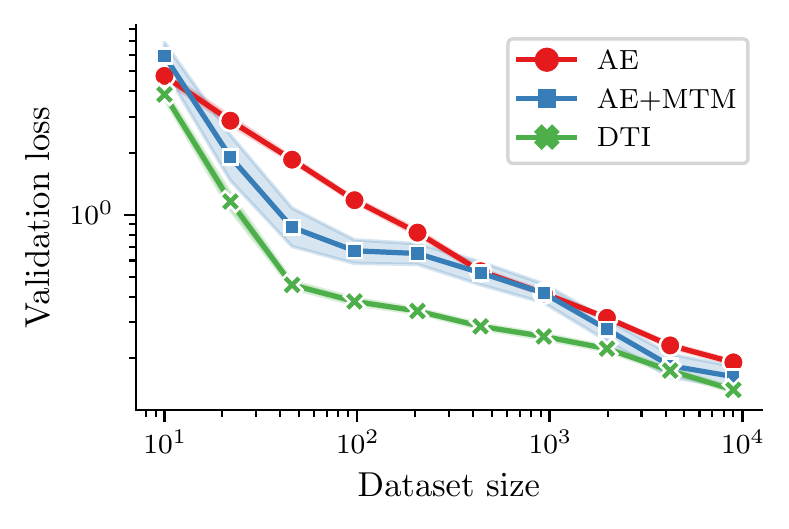}
    \includegraphics{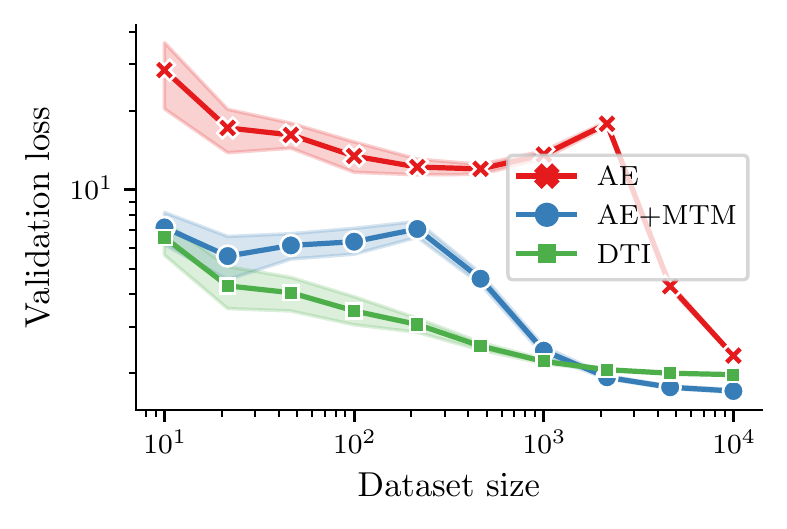}
    \caption{Validation loss versus dataset size for MNIST (top) and CIFAR (bottom). Datasets are generated by splitting the original validation set into a subset whose size is indicated by the ticks on the x-axis. Losses are then computed by training classifiers on an additional train split and evaluated on 10\% of the subset.}
    \label{fig:loss-data-curves}
\end{figure}
Because works like \citet{henaff2020data} and \citet{resnick2019probing} show that evaluation of learned representations by simple linear probes may not be sufficient, we adopt methods suggested in \citet{whitney2020evaluating} to evaluate the learned representations further. In particular, we evaluate the data efficiency of our representations, i.e. how well they perform on downstream tasks given validation sets of varying sizes, as detailed in section \ref{sec:evaluation}. We present the results in Fig.~\ref{fig:loss-data-curves}.

Firstly, note that given the original test set size ($10^4$), the results are consistent with our findings from the previous section. In MNIST, linear probes achieve the lowest loss on \textit{DTI} representations and in CIFAR, \textit{AE+MTM} leads to the lowest loss. Both in MNIST and CIFAR10, we find that these differences are significantly more pronounced in low data settings while this difference is most pronounced in CIFAR10. Here we find, using the least amount of data points ($10^1$), that both \textit{DTI} and \textit{AE+MTM} achieve an order of magnitude lower loss than \textit{AE}. Additionally, the results on CIFAR show that, while \textit{AE+MTM} achieves the lowest loss globally, \textit{DTI} achieves the lowest loss averaged over all dataset sizes. This indicates that $L^{DTI}$ is a good candidate loss to optimise in low data regimes, even for more complex datasets like CIFAR.

\begin{figure}[h]
    \centering
    \includegraphics{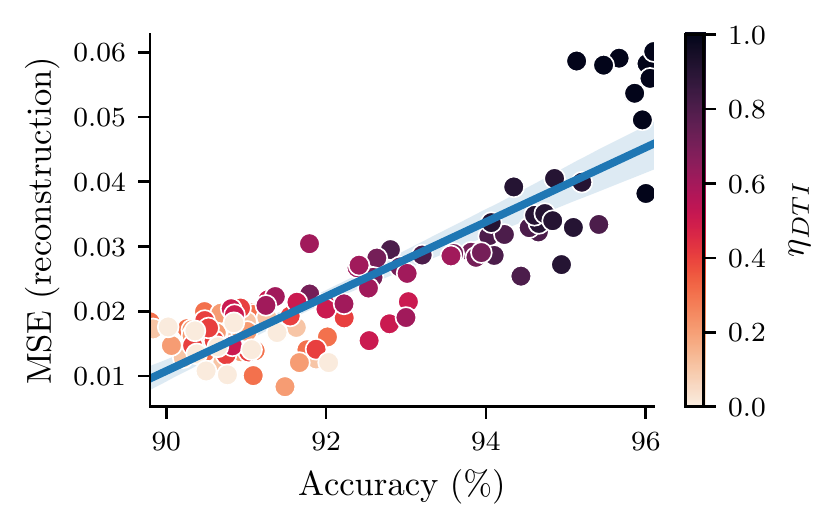}
    \caption{Reconstruction error versus downstream accuracy for different values of $\eta_{DTI}$ (with $\eta_{AE} = 1 - \eta_{DTI}).$ Blue line represents a ordinary least squares fit.}
    \label{fig:acc-vs-reconstruction-mnist}
\end{figure}

Next, we want to investigate how optimising for all our proposed losses (including $L^{AE}$) affects reconstruction performance and downstream accuracy in tandem. For this, we focus on the \textit{DTI} setting in MNIST. We interpolate $\eta_{AE}$ between 0 and 1 and set $\eta_{DTI} = 1 - \eta_{AE}$ and $\eta_{MTM} = \eta_{DTD} = 0$. We repeat the training of linear probes as in the previous sections using the full validation set. We present the results in Fig.~\ref{fig:acc-vs-reconstruction-mnist}.

The results show a positive correlation between accuracy and reconstruction error (\textit{p} < 0.05). This highlights a trade-off our method makes when optimising for representation performance. Our representation learning method in its current form relies both on reconstruction and on latent space transformation based on the input of other agents. We interpret these two forces as \textit{orthogonal}, because optimising for $L^{AE}$ favours a representation of the input that includes details not necessarily specific to the underlying class, whereas optimising for $L^{DTI}$ rewards representations which are particularly abstract and do not distinguish between individual samples of a particular class.

\subsection{Alignment of representations}
\begin{figure} 
    \centering
    \includegraphics{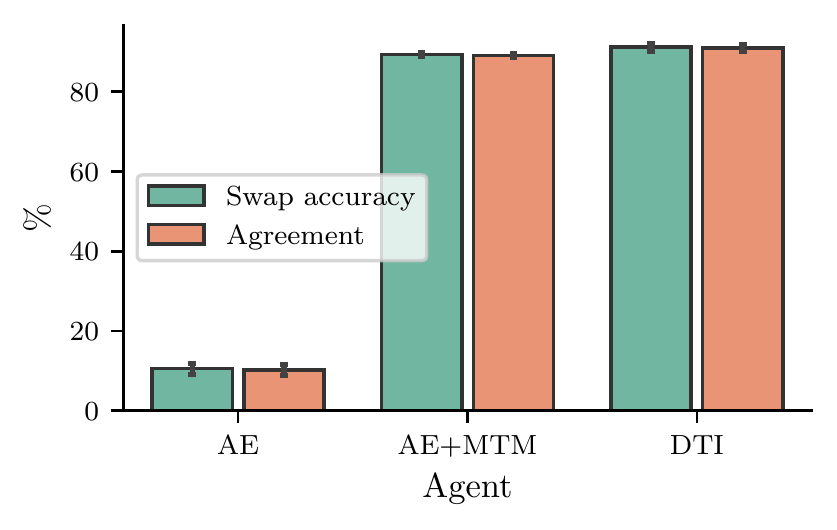}
    \includegraphics{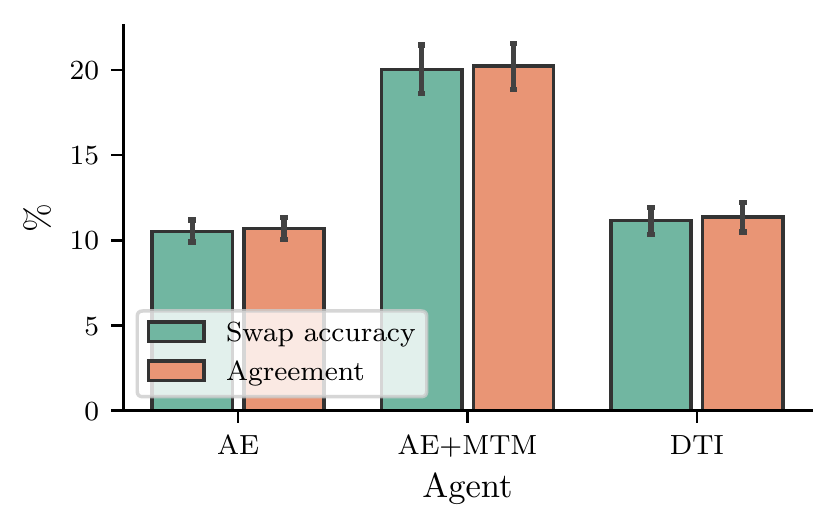}
    \caption{Swap accuracy and agreement for MNIST (top) and CIFAR (bottom).}
    \label{fig:swap_noise}
\end{figure}
Next, we examine the learned latent spaces for alignment.
To measure alignment, we use both swap accuracy and agreement, defined in section \ref{sec:methods}. We present the resulting plots in Fig.~\ref{fig:swap_noise}.

The data shows that for MNIST, both \textit{DTI} and \textit{AE+MTM} are able to achieve highly aligned latent spaces, whereas \textit{AE} alignment remains at chance level. The high alignment of \textit{DTI} and \textit{AE+MTM} coincides with the highclassification performance of both runs, indicating that successful alignment of latent spaces is important for learning effective latent spaces.

In CIFAR10, we observe that both \textit{AE} and \textit{DTI} remain close to chance level and only \textit{AE+MTM} achieves significant alignment, albeit much lower than in MNIST. We would like to highlight that in Fig.~\ref{fig:classification-accuracy-mnist} we show that \textit{AE+MTM} is the run with the best performance in CIFAR, further bolstering the hypothesis that successful latent space alignment is important for good representations.

\section{Discussions}
\label{sec:discussions}
In this work, we show that better data representations can be obtained using a population of agents optimising their respective latent space for communication while exploiting the property of subjectivity, which is inherent to multi-agent systems. We show that, using our proposed multi-agent losses, we can achieve an aligned protocol shared among all agents. Interestingly, we find that the best representations in our method also have highly aligned latent spaces. In addition, our results show that exploiting the inherent subjectivity of the systems seems to be crucial for better representations to arise. Thus, we conjecture that in the process of aligning their latent spaces, agents find more abstract data representations because the pressure to communicate under ambiguity selects for those representations that abstract away the view-dependent details of the underlying classes. Because we exploit the concept of shared attention in a multi-agent context, which can be interpreted as a weak supervision signal to agents, providing them access to latent representations of similar states, we call our method \emph{socially supervised}.

We also show that our representations are particularly \emph{data efficient}, by analysing performance on validation datasets of various sizes. Here we are able to show that optimising for $L^{DTI}$ is particularly effective in low-data regimes, yielding an order of magnitude loss decrease in CIFAR. Overall, we find that all of our proposed losses outperform the baselines consistently for almost all dataset sizes. Thus, because our representations need fewer data points to adapt to downstream tasks, they could potentially be employed in real-life settings where collecting large amounts of data is costly or dangerous.

Additionally, we want to highlight that under the assumption of no subjectivity and when only optimising $L^{DTI}$, our approach reduces to the approach introduced by  \citet{tieleman_shaping_2019} (see Related Work). Our results seem to contradict their finding that more efficient representations can be obtained without subjectivity.



Because our work is limited in scope, we can afford to investigate properties such as the interaction of our proposed losses deeply and with statistical rigor. Accordingly, we believe that the insights gained here should be taken to a more complex domain. One natural extension to our work would encompass the integration into a full multi-agent reinforcement learning loop. This is particularly interesting, because a shared context arises from multi-agent interaction in a natural way. When agents are in close vicinity, they most likely observe cohesive parts of their environments. It is this scenario that originally led us to the proposition that the benefits of data augmentation in self-supervised learning can emerge "for free" in multi-agent systems, a paradigm that we call \emph{Socially Supervised Representation Learning}. In such a setup, the improved representations could be used for policy learning or the building of effective world models. \citet{hafnerDreamControlLearning2020a}, for example, show that good representations are crucial for the success of model-based reinforcement learning \citep{wangBenchmarkingModelBasedReinforcement2019, moerlandModelbasedReinforcementLearning2020b}.

Lastly, we want to mention that while we use simple autoencoders here, our method is not bound to a particular representation learning method. Especially methods such as those described in \citet{grill2020bootstrap} could potentially employed in our proposed setup. While we discussed the similarity to \emph{contrastive learning} methods and how in \emph{SocSRL} we use perspectives as a substitute for augmented positive pairs, we think that negative pairs might also be of benefit. For example, agents might not only share messages about the currently observed object, but additionally a random sample from a (big) replay buffer, acting as a negative pair. We leave this investigation open for further research.

\begin{acks}
This research was partially funded by the French National Research Agency (\url{https://anr.fr/}, project ECOCURL, Grant ANR-20-CE23-0006) as well as the Inria Exploratory action ORIGINS (\url{https://www.inria.fr/en/origins}). This work also benefited from access to the HPC resources of IDRIS under the allocation 2020-[A0091011996] made by GENCI, using the Jean Zay supercomputer.
\end{acks}



\bibliographystyle{ACM-Reference-Format} 
\bibliography{main}


\end{document}